\documentclass{article}
\usepackage{spconf,amsmath,graphicx}
\usepackage{amsfonts}
\usepackage{dblfloatfix}
\usepackage{algorithm,algorithmic}

\newcommand{\R}{\mathbb{R}}

\newcommand{\orderof}[1]{\mathcal{O}\left(#1\right)}

\newcommand{\e}{e}

\newcommand{\vct}[1]{\boldsymbol{#1}}
\newcommand{\mtx}[1]{\boldsymbol{#1}}

\newcommand{\set}[1]{\mathcal{#1}}

\newcommand{\argmin}[1]{\underset{#1}{\operatorname{arg}\,\operatorname{min}}\;} 

\newcommand{\vc}{\vct{c}}
\newcommand{\vd}{\vct{d}}

\newcommand{\vq}{\vct{q}}

\newcommand{\vs}{\vct{s}}

\newcommand{\vv}{\vct{v}}

\newcommand{\vx}{\vct{x}}
\newcommand{\vy}{\vct{y}}
\newcommand{\vz}{\vct{z}}

\newcommand{\vepsilon}{\vct{\epsilon}}

\newcommand{\vmu}{\vct{\mu}}

\newcommand{\vsigma}{\vct{\sigma}}

\newcommand{\mPhi}{\mtx{\Phi}}

\newcommand{\mId}{{\bf I}}

\newcommand{\setB}{\set{B}}
\newcommand{\setC}{\set{C}}
\newcommand{\setD}{\set{D}}

\newcommand{\setS}{\set{S}}

\newtheorem{theorem}{Theorem}[section]

\newtheorem{lemma}[theorem]{Lemma}

\title{FAST COMPRESSIVE SENSING RECOVERY USING GENERATIVE MODELS WITH STRUCTURED LATENT VARIABLES}
\name{Shaojie Xu \qquad Sihan Zeng \qquad Justin Romberg
\thanks{\footnotesize This work was supported in part by Semiconductor Research Corporation (SRC).}
\thanks{\footnotesize Copyright 2019 IEEE. To appear in the IEEE 2019 International Conference on Acoustics, Speech, and Signal Processing (ICASSP 2019), scheduled for 12-17 May, 2019, in Brighton, United Kingdom. Personal use of this material is permitted. However, permission to reprint/republish this material for advertising or promotional purposes or for creating new collective works for resale or redistribution to servers or lists, or to reuse any copyrighted component of this work in other works, must be obtained from the IEEE. Contact: Manager, Copyrights and Permissions / IEEE Service Center / 445 Hoes Lane / P.O. Box 1331 / Piscataway, NJ 08855-1331, USA. Telephone: + Intl. 908-562-3966.}
}
\address{School of Electrical and Computer Engineering, Georgia Institute of Technology, Atlanta, GA, 30332\\
$\{$kyle.xu, szeng30$\}$@gatech.edu, $\{$jrom$\}$@ece.gatech.edu}
\begin{document}
%
\maketitle
\begin{abstract}
Deep learning models have significantly improved the visual quality and accuracy on compressive sensing recovery. In this paper, we propose an algorithm for signal reconstruction from compressed measurements with image priors captured by a generative model. We search and constrain on latent variable space to make the method stable when the number of compressed measurements is extremely limited. We show that, by exploiting certain structures of the latent variables, the proposed method produces improved reconstruction accuracy and preserves realistic and non-smooth features in the image. Our algorithm achieves high computation speed by projecting between the original signal space and the latent variable space in an alternating fashion.
\end{abstract}
%
\section{INTRODUCTION} \label{sec:intro}
In compressive sensing (CS), we seek to reconstruct a high-dimensional signal after observing a small number of linearly coded measurements. Mathematically, given a vector $\vx \in \R^N$, we obtain its linearly compressed representation $\vy$ in a low dimensional space $\R^M$ $(M \ll N)$ by applying $\vy = \mPhi \vx$, where $\mPhi$ is the compression matrix in $\R^{M\times N}$. In the case of useful compression, for two distinct signals $\vx_1$ and $\vx_2$ in the original space, their corresponding representations $\vy_1$ and $\vy_2$ in the compressed domain also need to be separate. As $\mPhi$ is an under-determined matrix with a null space, information can not be preserved for all vectors in $\R^N$. Fortunately, real-life signals often have prominent structures and only lie in some subspace of $\R^N$.

Traditional CS focuses on sparse signals. When a signal $\vx$ has (or approximately has) a k-sparse representation in some basis, the compression matrix $\mPhi$ can be constructed to preserve the distances between two signals in the compressed domain, a property formally known as the \emph{restricted isometry property} \cite{donoho2006compressed}. Common methods to construct $\mPhi$ include choosing each entry from an i.i.d. Gaussian \cite{baraniuk2008simple}, a fair Bernoulli, or an independent Sub-Gaussian distribution \cite{candes2008introduction}. Similar results were also established for signals on a smooth manifold. \cite{baraniuk2009random}

In order to retrieve the original signal from the compressed measurements, some type of prior knowledge needs to be assumed. When the class of signal is well-studied, the prior knowledge comes from years of experience. For example, the sparsity model is often used to recover compressively sensed natural images. Now with the power of machine learning (ML), we can extract prior knowledge efficiently from collected datasets. The powerful function approximation capability of deep neural networks also allow us to discover and represent more complicated signal structures.

In this paper, we propose a fast compressive sensing recovery algorithm using generative models with structured latent variables. The prior information of the signals is captured by a generative adversarial network (GAN). The stability of the recovery algorithm is improved when the GAN's latent variable space is well structured. Based on Alternating Direction Methods of Multipliers (ADMM), our algorithm achieves high reconstruction speed by alternatively projecting between the original signal space and the latent variable space, without involving gradient descent (GD). 

To the authors' knowledge, our work is the first to extend the ADMM-based CS recovery methods to GANs. Such an extension allows us to exploit the strong priors captured by GANs and significantly increases the recoverable compression ratio. Previous works on CS recovery using generative model either had limited model capacity \cite{chen2010compressive} or were slow to carry out \cite{bora2017compressed, kabkab2018task, shah2018solving} due to relying heavily on GD. We demonstrate a structured latent variable space in GANs plays an important role in fast and stable recovery. Our proposed algorithm achieves comparable performance with a notable speed-up compared to the gradient-based methods. Although We present our results by solving compressive sensing recovery problems, our model can be easily generalized to solve other inverse imaging problems.

\section{Related Work} \label{sec:related_work}
Signal recovery from the compressed measurements can be formulated as an optimization problem of the following form:
\begin{equation} \label{eq:CSrec}
\min_{\vx} \  F(\vx) + \lambda J(\vx)
\vspace{-1.5mm}
\end{equation}
The first term $F(\vx)$ is the fidelity term which controls how well a candidate signal $\vx$ matches the measurement $\vy$ in the compressed domain. A common choice of $F(\cdot)$ is to reflect the Euclidean distance in the compressed domain, such as $||\vy - \mPhi\vx||_2^2$. The second term $J(\cdot)$ is the property term which encodes the properties the signal of interest must satisfy. In the case of sparse signals, $J(\cdot)$ can be $||\vx||_0$ or the convex relaxation form $||\vx||_1$. When a dataset is available, one can ``learn" the properties of the signals using ML algorithms. Unsupervised learning methods such as Gaussian mixture model (GMM) \cite{chen2010compressive} and variational autoencoder (VAE) \cite{bora2017compressed} have been proposed. A scalar $\lambda$ controls the trade-off between the fidelity term and the property term.

Recovering sparse signals is a well-studied area. Algorithms such as orthogonal matching pursuit \cite{tropp2007signal}, linear programming \cite{donoho2006compressed}, least angle regression stagewise \cite{efron2004least}, soft thresholding \cite{parikh2014proximal}, and approximate message passing \cite{donoho2009message} were proposed to solve the variations of the optimization problem (\ref{eq:CSrec}).

When $J(\cdot)$ is learned directly from the data, it usually has a non-convex form. Although the global minimum is difficult to find, a local minimum may already result in satisfying results. One can also apply ADMM to (\ref{eq:CSrec}), which leads to iterative solving steps:
\begin{equation}
\begin{aligned}[b]
\vx^{(k+1)} &= \argmin{\vx} \  F(\vx)+\frac{\rho}{2}||\vx-\vs^{(k)}+\vmu^{(k)}||_2^2 \\
\vs^{(k+1)} &= \argmin{\vs} \  \lambda J(\vs)+\frac{\rho}{2}||\vx^{(k+1)}-\vs+\vmu^{(k)}||_2^2 \\
\vmu^{(k+1)} &= \vmu^{(k)}+\vx^{(k+1)}-\vs^{(k+1)}
\end{aligned}
\label{eq:FCSRG}
\end{equation}
ADMM introduces one auxiliary variable $\vs$ and one dual variable $\vmu$. Let $F(\cdot)$ take the form of $||\vy - \mPhi\vx||_2^2$, and the first update step in (\ref{eq:FCSRG}) can be easily computed by solving least squares. The second update step is a proximal operator of J, and can be understood as finding a signal $\vs$ that is close to the target signal $\vx^{(k+1)} + \vmu^{(k)}$ while satisfying the properties encoded in $J(\cdot)$. Such a formulation is often considered as a ``denoising" step and inspires many recent works on ``plug-and-play" methods. In these works, the second update step is replaced by state-of-the-art denoisers. Instead of explicitly learning $J(\cdot)$ to capture the statistics of the signals, denoisers are trained directly to mimic the behavior of the second update step with the properties of the signals embedded into the denoisers' design. Common denoiser choices include block-matching and 3D filtering (BM3D) \cite{dabov2009bm3d, metzler2015bm3d}, and feed-forward neural networks \cite{xie2012image, burger2012image}. Adversarial training was proposed in \cite{rick2017one, fan2017inner} to improve denoising and recovery. Discriminator were used to evaluate the denosing effect, however no generative model was trained to directly capture the statistics of the signal datasets. 

In the aforementioned ADMM framework, $\vx$ and $\vs$ are both in the original space $\R^N$ (with the ADMM underlying constraint $\vx=\vs$). When generative models are used to capture the signal statistics, latent variables $\vz \in \R^L$ are usually introduced. \cite{chen2010compressive} used a GMM to model a smooth manifold and searched $\vz^*$ in the latent space which has a closed form solution. \cite{bora2017compressed} used a VAE to learn a non-linear mapping $G_{gen}(\cdot)$ from $\vz \sim \mathcal{N}(\vmu,\,\vsigma^{2})$ to $\vx$ and applied GD to the optimization problem $\min\limits_{\vz} ||\vy - \mPhi G_{gen}(\vz)||_2^2 + \lambda ||\vz||_2^2$. Recent developments in GAN \cite{goodfellow2014generative,radford2015unsupervised} shed new lights on learning signal statistics. \cite{kabkab2018task} used a GAN to capture the image prior and applied GD to the same optimization problem as in \cite{bora2017compressed} for recovery. \cite{shah2018solving} proposed an algorithm that alternates between one step GD on the fidelity term and searching the latent variable space with the latter still achieved by GD. These gradient based methods suffer from high computational complexity and slow recovery speed. We seek to combine the fast computation from the ADMM-based methods and the strong prior-capture ability from the generative models to achieve fast CS recovery with ultra small number of measurements.

\section{ALGORITHMS} \label{sec:algorithms}
In the proposed algorithm, we formulate the CS recovery problem as searching $\vx$ and $\vz$ in the original signal space and in the latent variable space, respectively:
\begin{equation}
\begin{aligned}[b]
\min_{\vx, \vz} \quad & ||\vy - \mPhi\vx||_2^2 + \lambda H(\vz) \\
s.t. \quad & \vx = G_{gen}(\vz)
\end{aligned}
\label{eq:proposed_CS_rec}
\end{equation}
We denote $G_{gen}(\cdot)$ as the generative model and $H(\cdot)$ as the function that captures the property that the latent variable $\vz$ should satisfy. Even though the formulation above contains a non-linear equality constraint, we may still solve it by searching a stationary point of its augmented Lagrangian, which leads to the following ADMM-like update steps:
\begin{equation}
\small
\begin{aligned}[b]
\vx^{(k+1)} &= \left(\mPhi^T \mPhi+\rho\mId \right)^{-1}\left(
\mPhi^T\vy + \rho (G_{gen}(\vz^{(k)})-\vmu^{(k)})\right) \\
\vz^{(k+1)} &= \argmin{\vz} \  \lambda H(\vz)+\frac{\rho}{2}||\vx^{(k+1)}-G_{gen}(\vz)+\vmu^{(k)}||_2^2 \\
\vmu^{(k+1)} &= \vmu^{(k)}+\vx^{(k+1)}-G_{gen}(\vz^{(k+1)})
\end{aligned}
\label{eq:CSrecADMM}
\end{equation}

The property imposed by $H(\cdot)$ plays an important role in searching $\vz$ in the latent variable space. In classic GAN models such as the deep convolutional GAN (DCGAN) proposed in \cite{radford2015unsupervised}, $\vz$ is assumed to be drawn from a standard multivariate Gaussian distribution. Setting $H(\vz) = ||\vz||_2$ is a common practice to enforce such latent space structure. However, this latent variable space is still lack of interpretability and controllability, and how each latent dimension contributes to the generated signal is unclear \cite{chen2016infogan, chongxuan2017triple, deng2017structured}. The CS recovery performance bound provided in \cite{bora2017compressed} shows the number of required compressed measurements grows linearly with the latent variable dimension.

To enforce better structured latent variable space, we propose training the generative models with an InfoGAN setup \cite{chen2016infogan}, where each latent variable $\vz$ is split into a codeword $\vc$ and a ``random-noise-like" variable $\vv$. An InfoGAN is trained to not only minimize the usual GAN's loss function, but also maximize the mutual information between the codeword $\vc$ and the generated signal $G_{gen}(\vc,\vv)$. As a results, the codeword $\vc$ is able to control most of the semantic meaning in the generated signals, while $\vv$ only adds small variations to the results. This well-structured latent variable space helps CS recovery: When the number of compressed measurements is sufficient, both $\vc$ and $\vv$ may be inferred from the measurements, leading to more accurate reconstructions. When the number of compressed measurements is extremely limited, we can recover an approximate signal (within the small variation controlled by $\vv$) as long as $\vc$ can be inferred.

The x update step in (\ref{eq:CSrecADMM}) solves a least squares problem, while solving for the $\vz$ update is computationally expensive with GD. Similar to the ``plug-and-play" ADMM method, we propose to use a projector neural network $G_{proj}$ to learn the solution of this optimization problem. Notice that during each iteration, $\vx^{(k+1)}$ contains the noise introduced by the previous least square update.  We, therefore, propose to train $G_{proj}$ using randomly sampled latent variables and the noisy version of their generated samples. 

Alternatively, we can cascade the projector network and the generator network to form a network $G_{gen}(G_{proj}(\cdot))$ similar to an ``autoencoder". We then draw samples directly from the dataset, and train the ``autoencoder" to recover these samples from their noisy observations. $G_{gen}$ is fixed during this training process. When this method is used, our proposed algorithm is similar to a ``plug-and-play" ADMM model with a generative-model-based denoiser. Such a similarity may provide a better understanding about the convergence behavior of the proposed algorithm, while our previous derivation provides a better understanding about the importance of using well-structured latent variables. The complete version of the proposed algorithm is summarized in Algorithm \ref{algo:CSMLGEN}.

\begin{algorithm}[htb]
\small
\caption{Fast CS recovery using generative models (F-CSRG)} \label{algo:CSMLGEN}
\begin{algorithmic}
	\STATE Train a generative model $G_{gen}(\cdot)$ on the dataset.
	\STATE \textit{Method I:}
	\STATE \quad Randomly sample latent variable $\vz$ following its distribution.
	\STATE \quad Generate random noise $\vepsilon$ according to some distribution.
	\STATE \quad Construct noisy signal $\tilde{\vx}$ such that $\tilde{\vx} = G_{gen}(\vz) + \vepsilon$.
	\STATE \quad Train a projector network $G_{proj}(\cdot)$ that maps $\tilde{\vx}$ to $\vz$.
	\STATE \textit{Method II:}
	\STATE \quad Draw samples of $\vx$ from the training set.
	\STATE \quad Generate random noise $\vepsilon$ according to some distribution.
	\STATE \quad Construct noisy signals $\tilde{\vx}$ such that $\tilde{\vx} = \vx + \vepsilon$.
	\STATE \quad Train a projector network $G_{proj}(\cdot)$ such that $G_{gen}(G_{proj}(\cdot))$ maps $\tilde{\vx}$ to $\vx$.
	\STATE \quad\  $G_{gen}$ is fixed during training.
	\STATE
	\STATE \textbf{For signal recovery:}
  	\STATE Given a compression matrix $\mPhi$ and compressed measurements $\vy$.
 	\WHILE{Stopping criteria not met} 
 	\STATE $\vx^{(k+1)} = \left(\mPhi^T \mPhi+\rho\mId \right)^{-1}\left(
\mPhi^T\vy + \rho (G_{gen}(\vz^{(k)})-\vmu^{(k)})\right)$
    \STATE $\vz^{(k+1)} = G_{proj}(\vx^{(k+1)}+\vmu^{(k)})$
    \STATE $\vmu^{(k+1)} = \vmu^{(k)}+\vx^{(k+1)}-G_{gen}(\vz^{(k+1)})$
  	\ENDWHILE
\end{algorithmic}
\end{algorithm}

\begin{figure*}[htb]
\centering
\includegraphics[scale=.47]{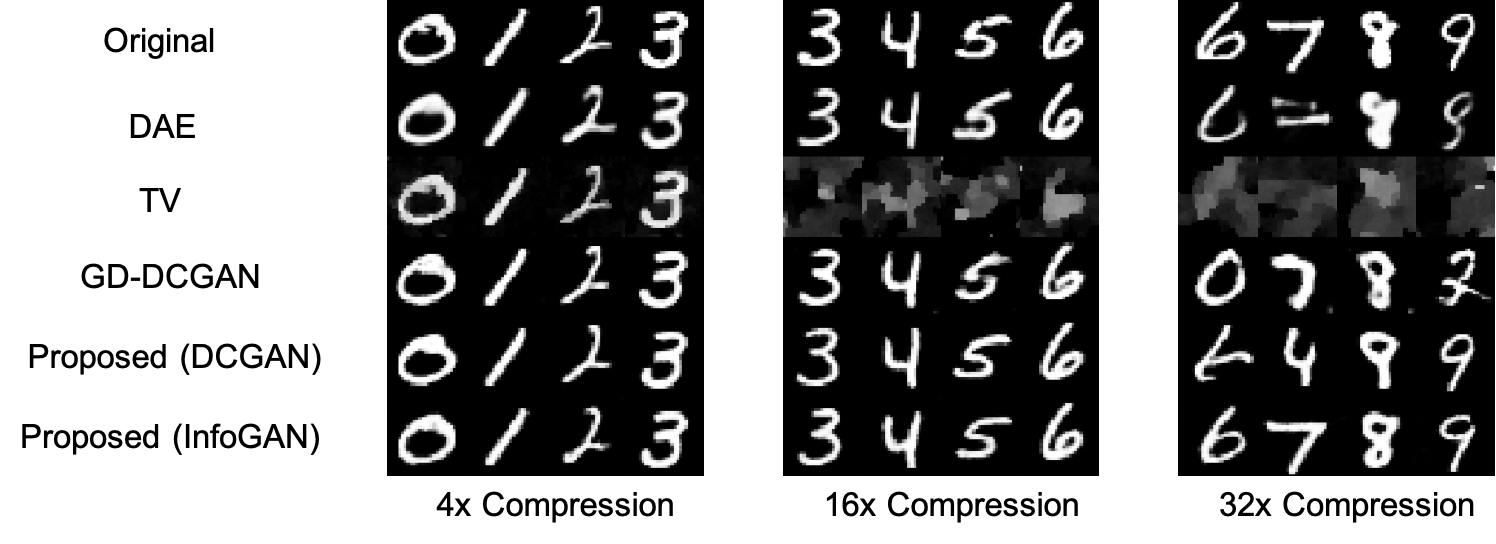}
\vspace{-.5cm}
\caption{\small Comparison of selected MNIST digits recovered by different algorithms.} \label{Fig:MLCS_algo_compare}
\end{figure*}

\begin{figure*}[htb]
\centering
\includegraphics[scale=.525]{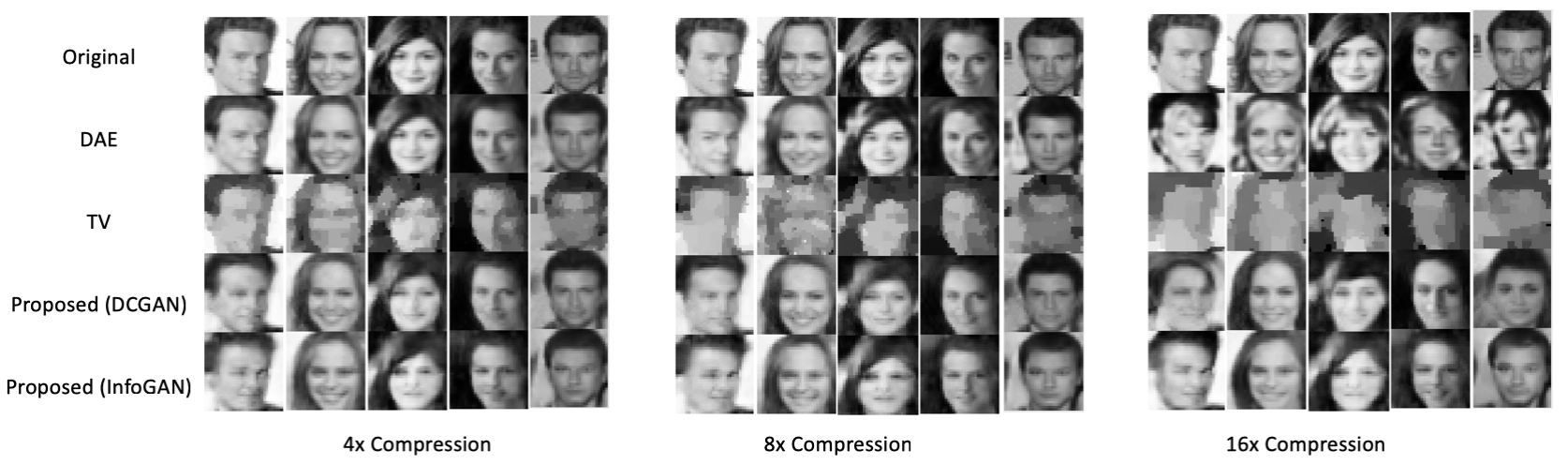}
\vspace{-.5cm}
\caption{\small Comparison of selected CelebA images recovered by different algorithms}
\label{Fig:CelebA_Results}
\vspace{-.5cm}
\end{figure*}

\section{Testing and results}
\label{sec:testing}

\subsection{MNIST Dataset}
We test the proposed algorithm using the MNIST digits dataset \cite{mnistdataset}. Selected recovered results are shown in Figure \ref{Fig:MLCS_algo_compare}. We use an i.i.d. Gaussian random matrix $\mPhi$ for compression. For comparison purposes, we also include the results of three baseline algorithms. The first algorithm is the ``plug-and-play" ADMM with a denoising autoencoder (DAE). As this denoiser is trained directly in the pixel domain (mapping $\tilde{\vx} =\vx+\vepsilon$ back to $\vx$), it fails to capture a strong prior knowledge about the digits, and starts to produce images with large artifacts as the compression rate goes to 32x. The second baseline algorithm uses the well-known total variance (TV) \cite{candes2006robust} as the regularizer. The third baseline algorithm uses DCGAN \cite{radford2015unsupervised} to capture the signal statistics, while the CS recovery is performed using gradient descent as proposed in \cite{bora2017compressed}. We use the same DCGAN's generator as $G_{gen}$ in our F-CSRG algorithm and train a neural network with fully connected layers (784-1024-512-256-100, ReLU activation) as $G_{proj}$. F-CSRG performs comparably with the gradient-descent-based algorithm, while only takes 1/20 of the computational time. When the number of compressed measurements is significantly reduced, both DCGAN-based models break down due to unstable projections caused by their less structured latent variable space. The best performance is achieved by incorporating InfoGAN (trained as suggested in \cite{chen2016infogan}) into our algorithm, demonstrating the benefit of having well-structured latent variables.

We test three fast recovery algorithms on MNIST digits' testing dataset and measure the Euclidean distance between the true images and the recovered images. The results are shown in Table \ref{Table:mnist_test}. As the compression ratio increases, our algorithm with InfoGAN outperforms the other algorithms. We also use the classification accuracy as another metric to assess the recovery quality. We train a convolutional neural network as the classifier which achieves $99.3\%$ accuracy on the original MNIST testing dataset. We then apply this MNIST classifier to the images reconstructed by the three algorithms. As shown in Table \ref{Table:mnist_test}, when the number of compressed measurements is extremely limited, our proposed InfoGAN algorithm achieves significantly higher accuracy than the other two methods, as a result of the InfoGAN's structured latent variable space.
\begin{table}[htb]
\small
\begin{center}
 \begin{tabular}{|c|c|c|c|}  
 \hline
 \shortstack{ \\ Compression \\ Ratio} & \shortstack{ \\ DAE \\ \  } & \shortstack{ \\ F-CSRG \\ with DCGAN} & \shortstack{ \\ F-CSRG \\ with InfoGAN}\\ 
 \hline
 4x & 2.20 / 98.2\% & 2.25 / 98.3\% & 2.68 / 97.7\% \\ 
 \hline
 8x & 2.54 / 97.8\% & 2.72 / 97.3\% & 3.06 / 97.2\% \\ 
 \hline
 16x & 3.23 / 94.8\% & 3.70 / 91.7\% & 3.79 / 93.8\% \\ 
 \hline
 32x & 5.13 / 73.5\% & 5.86 / 66.4\% & 5.37 / 77.4\% \\ 
 \hline
 64x & 7.33 / 41.8\% & 7.91 / 36.2\% & 7.43 / 48.0\% \\ 
 \hline
\end{tabular}
\end{center}
\caption{ Average reconstruction error (measurened as the Euclidean distance) and classification accuracy of the reconstructed digits on MNIST digits' testing dataset}
\label{Table:mnist_test}
\end{table}
\vspace{-.5cm}

\subsection{Celeb A Datasets}
We also test the proposed algorithm using the CelebA dataset \cite{liu2015faceattributes}. Each image is of dimension $32\times32$, cropped and downsampled from the original dataset. Generative models are trained based on the standard DCGAN and InfoGAN architectures as proposed in the original papers. For InfoGAN, we use 5 categorical codes (one-hot encoding with 10 classes), 5 continuous codes, and a randomness variable of length 128, producing a latent variable of length 183. We train a fully connected network (1024-512-256-183, ReLu activation) as $G_{proj}$. Bases on whether the desired codeword is categorical or continuous, we use softmax as the activation function or simply skip activation on the output layer. We test the proposed algorithm with 4x, 8x, and 16x compression. For comparison purposes, we also include the results of two baseline algorithms. Similar to the testing on the MNIST dataset, the first baseline algorithm uses the TV regularization, and the second one uses the ``plug-and-play" ADMM with a DAE trained directly in the pixel domain. Selected recovered results are shown in Figure \ref{Fig:CelebA_Results}.

As the compression rate increases, the quality of the images recovered by TV regularization degrades and few features on the face can be recovered. The recovery algorithm using a DAE produces comparable or sometimes even better reconstruction in the case of 4x and 8x compression, but becomes unstable under extremely high compression rate. In contrast, the proposed F-CSRG method is very stable against the drop in the number of compressed measurements. In addition, because generative model provides a strong image prior that assumes face images to have sharp features and not necessarily smooth everywhere, images reconstructed by the proposed algorithm preserve high-frequency contents of the original images. Out testing results also show that F-CSRG works better with an InfoGAN implementation than with a DCGAN. As we have discussed in previous sections, this improvement in recovered image quality comes from the more structured latent variable space produced by the InfoGAN's codeword.

\onecolumn
\section{Theoretical recovery guarantee} \label{sec:theory}
We begin this section by stating the main theoretical result that provides a guarantee on the recovery performance in the compressed domain with respect to the performance without compression.
\begin{theorem} \label{thm:recovery}
Assume a Lipschitz continuous mapping $G(\cdot,\cdot): \R^D\times\R^{L-D}\rightarrow\R^N$ that has Lipschitz constant $T$ and satisfies: $||G(\vc, \vv_1) - G(\vc, \vv_2)|| \leq \beta$, $\forall \vc \in \setB^D(r_c)$, $\vv_1, \vv_2 \in \setB^{L-D}(r_v)$, where $\setB^L(r)$ denotes a norm ball in $\R^L$ with radius $r$: $\setB^L(r) = \{ z | z\in\R^L, ||z||<r\}$ and $||\cdot||$ denotes the $L_2$ norm. Assume also a random matrix $\mPhi \in \R^{M \times N}$ whose entries are sampled from an i.i.d. Gaussian $\mathcal{N}(0, 1/M)$. The compressed measurements are obtained as $\vy = \mPhi\vx^* + w$, where $\vx^*$ is the desired signal and $w$ is an independent additive noise. Define $\bar{\vx} = \argmin{\vx\in G(\setB^D(r_c), \setB^{L-D}(r_v))} ||\vx^* - \vx||$ and $\hat{\vx} = \argmin{\vx\in G(\setB^D(r_c), \setB^{L-D}(r_v))} ||\vy - \mPhi\vx||$. 

For some $0 < \epsilon< \frac{1}{2}$, $0<\delta<1$, if $M \geq \orderof{\frac{D}{\epsilon^2}\log\frac{Tr_c}{\delta}}$, then with probability at least $1-\delta$:
\begin{equation} \label{eq:recovery1}
||\vx^*-\hat{\vx}|| \leq \frac{5+2\sqrt{\frac{N}{M}}-\epsilon}{1-\epsilon}||\vx^*-\bar{\vx}|| + \frac{6+2\sqrt{\frac{N}{M}}-2\epsilon}{1-\epsilon}\beta + \frac{2}{1-\epsilon}||w|| + \orderof{\delta} \ .
\end{equation}

Furthermore, if $M \geq \orderof{\frac{L}{\epsilon^2}\log\frac{T\sqrt{r_c^2+r_v^2}}{\delta}}$, then with probability at least $1-\delta$:
\begin{equation} \label{eq:recovery2}
||\vx^*-\hat{\vx}|| \leq \frac{5+2\sqrt{\frac{N}{M}}-\epsilon}{1-\epsilon}||\vx^*-\bar{\vx}|| + \frac{2}{1-\epsilon}||w|| + \orderof{\delta} \ .
\end{equation} 
\end{theorem}

The first requirement of this theorem is a Lipschitz continuous mapping $G(\cdot)$ from $\R^L$ to $\R^n$. Lipschitz continuity requires the difference between two outputs of the mapping generated by two inputs with bounded difference also be bounded up to a constant factor $T$:
\begin{equation}
||G(\vz_1) - G(\vz_2)|| \leq T ||\vz_1 - \vz_2|| \ .
\end{equation} 
We also require the input variable of the mapping to be divided into two variables each with a bounded norm. Their contribution to the output of the mapping are significantly different. The first vector $\vc$ controls the most variations, and once it is fixed, the other input vector $\vv$'s contribution to the output can be bounded by a small constant $\beta$. These requirements can be satisfied by well-trained InfoGAN naturally. When sampling from a bounded latent variable space, the outputs of a generative model fall on the manifold described by the training data. The input of an InfoGAN is also separated into a codeword and a randomness variable. The former controls the major variations in the generated results while the latter only adds some details.

The compressive sensing matrix $\mPhi$ is a random Gaussian matrix, whose entries are i.i.d. sampled. Our theoretical results can be extended to other types of random matrices by plugging in their corresponding concentration bounds. The compressed measurements are assumed to contain some independent additive noise $w$. We confine our search of the true solution in the output space of the generative model. Without compression, we define $\bar{\vx}$ as the true signal $\vx^*$'s projection onto the generative model's output space. $\bar{\vx}$ has the shortest euclidean distance to $\vx^*$ compared to all other points on the generative model's output space. This distance is unlikely to be completely zero due to the generative model's imperfect matching to the true signal space. This fundamental constraint present in the generative model itself without compressive sensing's involvement poses a lower bound on the recovery performance. With compressive sensing, we only have access to the compressed measurements $\vy$. In this case, we still conduct our search in the generative model's output space, trying to find the signal that, after compression, best matches $y$. Denote this signal as $\hat{\vx}$ and we seek to bound how far compressive sensing pushes this recovered signal away from the true signal $\vx^*$ with respect to the distance between $\bar{\vx}$ and $\vx^*$.

The performance bound we provide above contains two parts. In the first part, we consider the case when the number of compressed measurements are very limited. The required number of measurements in this case has a linear relationship with the dimension of the codeword space $D$. In section \ref{sec:testing}, we have shown that $D$ is often extremely small. The effect of compression mainly shows in two terms in the right hand side of (\ref{eq:recovery1}). The error caused by the generative model's approximation of the true signal's distribution is magnified by a factor of $\sqrt{\frac{N}{M}}$. Moreover, some details in the original signal are inevitably lost due to the small number of measurements. These details, bounded by $\beta$, is also magnified by the order of $\sqrt{\frac{N}{M}}$ and contributes to the recovery error bound. However as the number of measurements increases, the details in the original signal become more likely to be recovered. In the second part of the theorem, when the number of measurements satisfies some linear relationship with the dimension of the whole latent variable space $L$, the term involving $\beta$ is dropped from the right hand side in (\ref{eq:recovery2}). This bound has been previously derived in \cite{bora2017compressed}. We extend the previous work to generative models whose latent variable space is well-structured and provided recovery guarantees adaptive to the number of measurements.

\subsection{Proof of the main theorem}
Like many previous theoretical works in compressive sensing, our proof starts from the Johnson-Lindenstrauss (JL) lemma, which states the length preserving property of applying random projection on a finite set of vectors.
\begin{lemma}[JL lemma with i.i.d. Gaussian matrix]
For a finite set $\setS$ of $Q$ vectors ($|\setS| = Q$) in $\R^N$, a random matrix $\mPhi \in \R^{M  \times N}$ whose entries are sampled from an i.i.d. Gaussian $\mathcal{N}(0, 1/M)$, and some $0 < \epsilon < \frac{1}{2}$, we have with probability at least $1 - 2Q\e^{-\frac{\epsilon^2M}{8}}$, $\forall \vx \in \setS$:
\begin{equation}
(1-\epsilon)||\vx|| \leq ||\mPhi\vx|| \leq (1+\epsilon)||\vx|| \ .
\end{equation}
In other words, for some $0 < \epsilon < \frac{1}{2}$ and probability $0 < \delta < 1$, $(1-\epsilon)||\vx|| \leq ||\mPhi\vx|| \leq (1+\epsilon)||\vx||$ holds with probability at least $1-\delta$ as long as:
\begin{equation}
M \geq \frac{8}{\epsilon^2}\log\frac{2Q}{\delta} \ .
\end{equation}
\end{lemma}

\textbf{\textit{Pairwise distance preserving in a covering set:}} Now consider a Lipschitz mapping $G(\cdot): \R^L\rightarrow\R^N$ with Lipschitz constant $T$, and a norm ball $\setB^L(r) = \{ z | z\in\R^L, ||z||<r\}$. We can construct an $\frac{\eta}{T}$-covering for $\setB^L(r)$, denoted as $\setC^L(\frac{\eta}{T})$. Then by the definition and covering and Lipschitz continuity $\setC^N(\eta)=G(\setC^L(\frac{\eta}{T}))$ is an $\eta$-covering for $G(\setB^L(r))$. Applying the classic bound on the covering number, we have:
\begin{equation}
|\setC^N(\eta)| = |\setC^L(\frac{\eta}{T})| \leq (\frac{3Tr}{\eta})^L \ .
\end{equation}
To prove the pairwise distance preserving property, we construct a new set $\setD^N(\eta) = \{\vd|\vd=\vx_1 - \vx_2,\ \vx_1,\vx_2\in\setC^N(\eta)\}$. Then by the way we construct $\setD^N(\eta)$, we have:
\begin{equation}
|\setD^N(\eta)| = \binom{|\setC^N(\eta)|}{2} \leq \frac{1}{2} |\setC^N(\eta)|^2 \leq \frac{1}{2} (\frac{3Tr}{\eta})^{2L} \ .
\end{equation}
 
Plugging the upper bound as Q in the JL Lemma, we derive that as long as
\begin{equation}
M \geq \frac{8}{\epsilon^2}\log(\frac{1}{\delta}(\frac{3Tr}{\eta})^{2L}) \ ,
\end{equation}
$(1-\epsilon)||\vx_1-\vx_2|| < ||\mPhi(\vx_1-\vx_2)|| < (1+\epsilon)||\vx_1-\vx_2||$ holds with probability at least $1-\delta$, for all $\vx_1,\vx_2\in\setC^N(\eta)$. And by setting $\eta = 3\delta^\frac{2L-1}{2L} =\orderof{\delta}$, we have:
\begin{equation}
M \geq \frac{16L}{\epsilon^2}\log\frac{Tr}{\delta} \ ,
\end{equation}
is sufficient to have $(1-\epsilon)||\vx_1-\vx_2|| < ||\mPhi(\vx_1-\vx_2)|| < (1+\epsilon)||\vx_1-\vx_2||$ hold with probability at least $1-\delta$, for all $\vx_1,\vx_2\in\setC^N(\orderof{\delta})$.

\textbf{\textit{RIP for all signals in the output space of a generative model:}} To generalize the distance preserving property to all $\vx_1,\vx_2\in G(\setB^L(r))$, we apply a technique named "chaining". First, we construct a chain of coverings for $G(\setB^L(r))$ with increasing radius. Then for every $\vx \in G(\setB^L(r))$, we can find a chain of vectors $\vq_F \in \setC_F^N(\frac{\eta}{2^F})$, $\vq_{F-1} \in \setC_{F-1}^N(\frac{\eta}{2^{F-1}})$, $\ldots$, $\vq_0 \in \setC_0^N(\eta)$ that satisfies $||\vx-\vq_F|| \leq \frac{\eta}{2^F}$ and $||\vq_{f+1}-\vq_f|| \leq \frac{\eta}{2^f}, \forall f=0,1,\ldots, F-1$. The construction starts from finding $\vq_F$ and then moves down along the chain. By the definition of covering, $\vq_f$ must exists given $\vq_{f+1}$. Even though the chaining vectors change with $\vx$, the coverings are fixed. In this way, the bound we derive later will hold for all $\vx$ simultaneously. The chaining processing also implies:
\begin{equation} \label{eq:chaining_sum_dist}
\begin{split}
||\vx - \vq_0|| &= ||(\vx-\vq_F)+(\vq_F-\vq_{F-1})+\ldots+(\vq_1-\vq_0)|| \\
&\leq ||\vx-\vq_F||+||\vq_F-\vq_{F-1}||+\ldots+||\vq_1-\vq_0|| \\ 
&\leq \sum_{f=0}^{F} \frac{\eta}{2^f} \leq 2\eta \ .
\end{split}
\end{equation}

With the constructed chain, the compressed distance from $\vx$ to a point in $\setC_0^N(\eta)$ covering can be bounded as:
\begin{equation} \label{eq:chaining_sum_dist_c}
\begin{split}
||\mPhi(\vx - \vq_0)|| &= ||\mPhi(\vx-\vq_F)+\mPhi(\vq_F-\vq_{F-1})+\ldots+\mPhi(\vq_1-\vq_0)||\\
&\leq ||\mPhi(\vx-\vq_F)||+||\mPhi(\vq_F-\vq_{F-1})||+\ldots+||\mPhi(\vq_1-\vq_0)|| \ .
\end{split}
\end{equation}
Construct new sets $\setD^N_f = \{\vd|\vd=\vq - \vq',\ \vq\in\setC^N_{f+1}(\frac{\eta}{2^{f+1}}),\ \vq'\in\setC^N_{f}(\frac{\eta}{2^{f}})\}$. Then by the way we construct $\setD^N_f$, we have:
\begin{equation} \label{eq:chaining_set_size}
|\setD^N| \leq |\setC^N_{f+1}(\frac{\eta}{2^{f+1}})|\ |\setC^N_{f+1}(\frac{\eta}{2^{f}})| \leq |\setC^N_{f+1}(\frac{\eta}{2^{f+1}})|^2 \leq (\frac{6Tr}{\eta})^{2L} 4^{fL} \ .
\end{equation}

To bound each term in (\ref{eq:chaining_sum_dist_c}), we apply a variation of the JL lemma:
\begin{lemma} \label{lm:JL_variation}
For a finite set $\setS$ of $Q$ vectors ($|\setS| = Q$) in $\R^N$, a random matrix $\mPhi \in \R^{M  \times N}$ whose entries are sampled from an i.i.d. Gaussian $\mathcal{N}(0, 1/M)$, and some $\epsilon > 0$, we have with probability at least $1 - Q((1+\epsilon)\e^{-\epsilon})^{M/2}$, $\forall \vx \in \setS$:
\begin{equation}
||\mPhi\vx|| \leq (1+\epsilon)||\vx|| \ .
\end{equation}
\end{lemma}
Plugging (\ref{eq:chaining_set_size}) as Q in Lemma \ref{lm:JL_variation}, we derive that with probability at least $1-\delta_f = 1 - (\frac{6Tr}{\eta})^{2L} 4^{fL}((1+\epsilon_f)\e^{-\epsilon_f})^{\frac{M}{2}}$,
\begin{equation}
||\mPhi(\vq_{f+1}-\vq_f)|| \leq (1+\epsilon_f)||\vq_{f+1}-\vq_f|| \leq (1+\epsilon_f)\frac{\eta}{2^f} \ .
\end{equation}

By applying a union bound, we have with probability at least $1-\sum\limits_{f=0}^{F-1} \delta_f$:
\begin{equation}  \label{eq:chaining_dist_sum}
\sum_{f=0}^{F-1}||\mPhi(\vq_{f+1}-\vq_f)|| \leq \sum_{f=0}^{F-1} (1+\epsilon_f)\frac{\eta}{2^f} \ .
\end{equation}
In order for this bound to be useful, we require that $\sum\limits_{f=0}^{F-1} \delta_f$ and $\sum\limits_{f=0}^{F-1} (1+\epsilon_f)\frac{\eta}{2^f}$ to both converge as $F\to\infty$. Let's consider the following sequence of $\epsilon_f$:
\begin{equation}\label{eq:epsilon_seq}
\epsilon_f = 
\begin{cases} 
\epsilon_0(1+f),\quad f \geq \left(\frac{1}{2\epsilon_0}\right)^2 - 1 \\
\epsilon_0\sqrt{1+f},\quad f < \left(\frac{1}{2\epsilon_0}\right)^2 - 1 \ ,
\end{cases}
\end{equation}
with some $\epsilon_0 < \frac{1}{2}$. Denote the splitting threshold as $\bar{f}$: $\bar{f} = \left(\frac{1}{2\epsilon_0}\right)^2 - 1$. Then (\ref{eq:chaining_dist_sum}) is in the order of $\eta$:
\begin{equation}
\begin{split}
\sum_{f=0}^\infty (1+\epsilon_f)\frac{\eta}{2^f} &= \sum_{f=0}^{\bar{f}} (1+\epsilon_0\sqrt{1+f})\frac{\eta}{2^f} + \sum_{f=\bar{f}}^\infty (1+\epsilon_0(1+f))\frac{\eta}{2^f} \\
&\leq \sum_{f=0}^\infty (1+\epsilon_0(1+f))\frac{\eta}{2^f} \\
&= (2+4\epsilon_0)\eta \ .
\end{split}
\end{equation}

To bound the probability term $1-\sum\limits_{f}\delta_f$, we have:
\begin{equation}
\sum_{f=0}^\infty \delta_f = \sum_{f=0}^\infty (\frac{6Tr}{\eta})^{2L} 4^{fL}((1+\epsilon_f)\e^{-\epsilon_f})^{\frac{M}{2}} \ .
\end{equation}
Notice when $\epsilon_f < \frac{1}{2}$, $(1+\epsilon_f)\e^{-\epsilon_f} < \e^{-\frac{1}{4}\epsilon_f^2}$, and when $\epsilon_f \geq \frac{1}{2}$, $(1+\epsilon_f)\e^{-\epsilon_f} > \e^{-\frac{1}{14}\epsilon_f}$. In our proposed $\epsilon$ sequence (\ref{eq:epsilon_seq}), $\epsilon_f \geq \frac{1}{2}$ if and only if $f\geq\bar{f}$. Therefore, we bound the probability term by:
\begin{equation}
\begin{split}
\sum_{f=0}^\infty \delta_f &= (\frac{6Tr}{\eta})^{2L}\left(\sum_{f=0}^{\bar{f}} 4^{fL}((1+\epsilon_f)\e^{-\epsilon_f})^{\frac{M}{2}} + \sum_{f=\bar{f}}^\infty4^{fL}((1+\epsilon_f)\e^{-\epsilon_f})^{\frac{M}{2}}\right)\\
&< (\frac{6Tr}{\eta})^{2L}\left(\sum_{f=0}^{\bar{f}} 4^{fL}\e^{-\frac{M}{8}\epsilon_f^2} + \sum_{f=\bar{f}}^\infty 4^{fL}\e^{-\frac{M}{14}\epsilon_f}\right)\\
&= (\frac{6Tr}{\eta})^{2L}\left( \e^{-\frac{M}{8}\epsilon_0^2}\sum_{f=0}^{\bar{f}} \left(4^{L}\e^{-\frac{M}{8}\epsilon_0^2}\right)^f + \e^{-\frac{M}{14}\epsilon_0}\sum_{f=\bar{f}}^\infty \left(4^{L}\e^{-\frac{M}{14}\epsilon_0}\right)^f \right) \ . \\
\end{split}
\end{equation}
Given $\epsilon_0 < \frac{1}{2}$, we have $\e^{-\frac{M}{14}\epsilon_0} < \e^{-\frac{M}{8}\epsilon_0^2}$. By denoting $\delta$ as the upper bound of the sum of $\delta_f$, we have:
\begin{equation}
\sum_{f=0}^\infty \delta_f < 
(\frac{6Tr}{\eta})^{2L} \e^{-\frac{M}{8}\epsilon_0^2}\sum_{f=0}^{\infty} \left(4^{L}\e^{-\frac{M}{8}\epsilon_0^2}\right)^f  \leq \delta \ . \\
\end{equation}
Then by setting $\eta = \delta^{\frac{2L-1}{2L}} = \orderof{\delta}$, the inequality above leads to:
\begin{equation}
M \geq \frac{8}{\epsilon_0^2}\log\left((\frac{6Tr}{\delta})^{2L} + 2^{2L}\right) \ .
\end{equation}
Since $||\mPhi(\vx-\vq_F)||\to0$ when $F\to\infty$, we conclude that with $M \geq \orderof{\frac{L}{\epsilon^2}\log\frac{Tr}{\delta}}$, and probability at least $1-\delta$, for any $\vx \in G(\setB^k(r)),\ \exists \vq\in\setC^N(\orderof{\delta})$, such at
\begin{equation} \label{eq:chaining_ieq}
||\mPhi(\vx - \vq)|| \leq \orderof{\delta} \ .
\end{equation}

Now, for every pair of vectors $\vx_1, \vx_2 \in G(\setB^L(r))$, we apply chaining to find two corresponding vectors $\vq_1, \vq_2$ such that $\vq_1,\vq_2 \in \setC^N(\orderof{\delta})$, and according to (\ref{eq:chaining_sum_dist}), $||\vx_1 - \vq_1|| \leq \orderof{\delta}$, $||\vx_2 - \vq_2|| \leq \orderof{\delta}$. We then apply the triangular inequality:
\begin{equation} \label{eq:pairwise_dist_ieq}
||\mPhi(\vx_1-\vx_2)|| \leq ||\mPhi(\vq_1-\vq_2)|| + ||\mPhi(\vx_1-\vq_1)|| + ||\mPhi(\vq_2-\vx_2)|| \ .
\end{equation}
The first term on the right hand side is the distance between two vectors on the covering set. We have shown that with $M \geq \frac{16L}{\epsilon^2}\log\frac{Tr}{\delta}$ and probability at least $1-\delta$, we have:
\begin{equation} \label{eq:center_dist_ieq}
\begin{split}
||\mPhi(\vq_1-\vq_2)|| & \leq (1+\epsilon)||\vq_1-\vq_2|| = (1+\epsilon)||\vq_1-\vq_2-\vx_1+\vx_1-\vx_2+\vx_2|| \\
& \leq (1+\epsilon)||\vx_1-\vx_2|| + (1+\epsilon)||\vx_1-\vq_1||+ (1+\epsilon)||\vx_2-\vq_2|| \\
& \leq (1+\epsilon)||\vx_1-\vx_2|| + (1+\epsilon)\orderof{\delta} \ .
\end{split}
\end{equation}

We can now obtain the pairwise distance preserving property by plugging (\ref{eq:center_dist_ieq}) and (\ref{eq:chaining_ieq}) into (\ref{eq:pairwise_dist_ieq}) and apply the union bound on the probabilities. We have only shown the derivation of the upper bound. The lower bound can be derived using similar techniques. The distance preserving property is formally stated in Lemma \ref{lm:lipschitz_rip}.
\begin{lemma}\label{lm:lipschitz_rip}
Assume a Lipschitz continuous mapping $G(\cdot): \R^L\rightarrow\R^N$ with Lipschitz constant $T$ and a norm ball $\setB^L(r) = \{ z | z\in\R^L, ||z||<r\}$ where $||\cdot||$ denotes the $L_2$ norm. Assume also a random matrix $\mPhi \in \R^{M \times N}$ whose entries are sampled from an i.i.d. Gaussian $\mathcal{N}(0, 1/M)$. Given some $0 <\epsilon<\frac{1}{2}$, $0<\delta<1$, if $M \geq \orderof{\frac{L}{\epsilon^2}\log\frac{Tr}{\delta}}$, then with probability at least $1-\delta$:
\begin{equation}
(1-\epsilon)||\vx_1-\vx_2|| - \orderof{\delta} \leq ||\mPhi(\vx_1-\vx_2)|| \leq (1+\epsilon)||\vx_1-\vx_2|| + \orderof{\delta}
\end{equation}
holds $\forall \vx_1,\vx_2 \in G(\setB^L(r)).$
\end{lemma}

\textbf{\textit{Compressive sensing recovery guarantee using a generative model:}} When the compressed measurements $\vy = \mPhi\vx^* + w$ are collected, we search in the subspace generated by $G(\setB^L(r))$ rather than $\R^n$.  Define $\bar{\vx}$ as the true signal's projection onto the output space of the generative model when no compression is presented:
\begin{equation}
\bar{\vx} = \argmin{\vx\in G(\setB^L(r))} ||\vx^* - \vx|| \ .
\end{equation}
When given only the compressed measurements $\vy$, we search the generative model's output space to find a signal that, after compression, best matches the compressed observations. Define this solution as $\hat{\vx}$:
\begin{equation}
\hat{\vx} = \argmin{\vx\in G(\setB^L(r))} ||\vy - \mPhi\vx|| \ .
\end{equation}

To show the guarantee of the compressive sensing recovery is to show that $||\vx^* - \hat{\vx}||$ is comparable to $||\vx^* - \bar{\vx}||$, essentially demonstrating that the compression has very small influence on the recovery. We start the proof by applying the triangular inequality:
\begin{equation}
||\vx^*-\hat{\vx}|| \leq ||\vx^*-\bar{\vx}|| + ||\bar{\vx}-\hat{\vx}|| \ .
\end{equation}
Since $\hat{\vx}, \bar{\vx} \in G(\setB^k(r))$, Lemma \ref{lm:lipschitz_rip} indicates that with $M \geq \orderof{\frac{L}{\epsilon^2}\log\frac{Tr}{\delta}}$ and probability at least $1 - \delta$:
\begin{equation} \label{eq:recovery_ieq}
\begin{split}
||\bar{\vx}-\hat{\vx}|| &\leq \frac{||\mPhi(\bar{\vx}-\hat{\vx})||+\orderof{\delta}}{1-\epsilon}\\
&\leq \frac{||\vy-\mPhi\bar{\vx}||+||\vy-\mPhi\hat{\vx}||+\orderof{\delta}}{1-\epsilon}\\
&\leq \frac{2||\vy-\mPhi\bar{\vx}||+\orderof{\delta}}{1-\epsilon}\\
&\leq \frac{2||\mPhi(\vx^*-\bar{\vx})||+2||w||+\orderof{\delta}}{1-\epsilon} \ .
\end{split}
\end{equation}

To bound the term $||\mPhi(\vx^*-\bar{\vx})||$, we apply the results of the random matrix \cite{davidson2001local, vershynin2010introduction, bandeira2016sharp}:
\begin{lemma} \label{lm:l2_norm_rand_matrix}
For a random matrix $\mPhi \in \R^{M \times N}$ whose each entry is sampled from an i.i.d. Gaussian $\mathcal{N}(0, 1/M)$, we have with probability at least $1 - 2\e^{-\frac{M}{2}}$, $\forall \vx \in \R^N$:
\begin{equation}
||\mPhi\vx|| \leq (2+\sqrt{\frac{N}{M}})||\vx|| \ .
\end{equation}
\end{lemma}
That is, with probability at least $1-\delta'=1-2\e^{-\frac{M}{2}}$:
\begin{equation} \label{eq:rand_matrix_ieq}
||\mPhi(\vx^*-\bar{\vx})|| \leq (2+\sqrt{\frac{N}{M}})||\vx^*-\bar{\vx}|| \ .
\end{equation}
It can be shown that with $M \geq \orderof{\frac{L}{\epsilon^2}\log\frac{Tr}{\delta}}$, a small $\epsilon$, and some reasonable values for r, T, L, $\delta' < \delta$. By plugging (\ref{eq:rand_matrix_ieq}) into (\ref{eq:recovery_ieq}) and combining the two probability with a union bound, we arrive at the compressive sensing recovery guarantee using a generative model:
\begin{lemma} \label{lm:recovery2}
Assume a Lipschitz continuous mapping $G(\cdot): \R^L\rightarrow\R^N$ with Lipschitz constant $T$ and a norm ball $\setB^L(r) = \{ z | z\in\R^L, ||z||<r\}$ where $||\cdot||$ denotes the $L_2$ norm. Assume also a random matrix $\mPhi \in \R^{M \times N}$ whose entries are sampled from an i.i.d. Gaussian $\mathcal{N}(0, 1/M)$. The compressed measurements are obtained as $\vy = \mPhi\vx^* + w$, where $\vx^*$ is the desired signal and $w$ is an independent additive noise. Define $\bar{\vx} = \argmin{\vx\in G(\setB^L(\vz))} ||\vx^* - \vx||$ and $\hat{\vx} = \argmin{\vx\in G(\setB^L(\vz))} ||\vy - \mPhi\vx||$. 

If $M \geq \orderof{\frac{L}{\epsilon^2}\log\frac{Tr}{\delta}}$, then with probability at least $1-\delta$:
\begin{equation}
||\vx^*-\hat{\vx}|| \leq \frac{5+2\sqrt{\frac{N}{M}}-\epsilon}{1-\epsilon}||\vx^*-\bar{\vx}|| + \frac{2}{1-\epsilon}||w|| + \orderof{\delta} \ .
\end{equation}
\end{lemma}

\textbf{\textit{Compressive sensing recovery guarantee using a generative model with structured latent variable space:}} We now impose some structures in the input space of the generative model. Assume a Lipschitz continuous mapping $G(\cdot,\cdot): \R^D\times\R^{L-D}\rightarrow\R^N$ with Lipschitz constant $T$ further satisfies: $||G(\vc, \vv_1) - G(\vc, \vv_2)|| \leq \beta,\ \forall \vc\in \setB^D(r_c), \vv_1, \vv_2 \in \setB^{L-D}(r_v)$. The input space of the generative model is divided into two subspaces. The first input vector $\vc \in \R^D$ controls the most variations, and once it is fixed the magnitude of the other input vector $\vv \in R^{L-D}$'s contribution to the output is bounded by a small constant $\beta$. 

As in the previous proof, we start from the step of constructing a covering set $\setC^N(\eta)$ for $G(\setB(r_c), \vv_0)$, where $\vv_0$ is arbitrary and fixed. As the generated vectors are fully controlled by $\vc$ in this way, we have:
\begin{equation}
|\setC^n(\eta)| \leq (\frac{3Tr_c}{\eta})^D \ .
\end{equation}
By the assumption of the generative model, for any two signals generated by the model $\vx_1 = G(\vc_1, \vv_1)$, $\vx_2 = G(\vc_2, \vv_2)$, we can find $\vx_{c1} = G(\vc_1, \vv_0)$ and $\vx_{c2} = G(\vc_2, \vv_0)$ such that:
\begin{equation}
||\vx_1 - \vx_{c1}|| \leq \beta,\ ||\vx_2 - \vx_{c2}|| \leq \beta \ .
\end{equation}
Applying the triangular inequality, the pairwise distance after compression can be bounded by:
\begin{equation} \label{eq:ls_triangular}
||\mPhi(\vx_1-\vx_2)|| \leq ||\mPhi(\vx_{c1}-\vx_{c2})|| + ||\mPhi(\vx_1-\vx_{c1})|| + ||\mPhi(\vx_2-\vx_{c2})|| \ .
\end{equation}

As $\vx_{c1}, \vx_{c2} \in G(\setB(r_c), \vv_0)$, we apply Lemma \ref{lm:lipschitz_rip}. With $M \geq \orderof{\frac{D}{\epsilon^2}\log\frac{Tr_c}{\delta}}$ and probability at least $1-\delta$:
\begin{equation}
||\mPhi(\vx_{c1}-\vx_{c2})|| \leq (1+\epsilon)||\vx_{c1}-\vx_{c2}|| + \orderof{\delta} \ . 
\end{equation}
For the second and third terms in (\ref{eq:ls_triangular}), we use Lemma \ref{lm:l2_norm_rand_matrix}. With probability as least $1-\delta' = 1-2\e^{-\frac{M}{2}}$:
\begin{equation}
||\mPhi(\vx_1-\vx_{c1})|| \leq (2+\sqrt{\frac{N}{M}})||\vx_1-\vx_{c1}|| \leq (2+\sqrt{\frac{N}{M}})\beta \ .
\end{equation}
And with probability as least $1-\delta'' = 1-2\e^{-\frac{M}{2}}$:
\begin{equation}
||\mPhi(\vx_2-\vx_{c2})|| \leq (2+\sqrt{\frac{N}{M}})||\vx_2-\vx_{c2}|| \leq (2+\sqrt{\frac{N}{M}})\beta \ .
\end{equation}
With $M \geq \orderof{\frac{D}{\epsilon^2}\log\frac{Tr_c}{\delta}}$, a small $\epsilon$, and some reasonable values for r, T, L, we can show that $\delta' < \delta$ and $\delta'' < \delta$. 

Now we can combine the the upper bound of each term in (\ref{eq:ls_triangular}) and conclude that with $M \geq \orderof{\frac{D}{\epsilon^2}\log\frac{Tr_c}{\delta}}$ and probability as least $1-3\delta$:
\begin{equation}
\begin{split}
||\mPhi(\vx_1-\vx_2)|| &\leq (1+\epsilon)||\vx_{c1}-\vx_{c2}|| + (4+2\sqrt{\frac{N}{M}})\beta + \orderof{\delta} \\
&\leq (1+\epsilon)||\vx_1-\vx_2|| + (1+\epsilon)||\vx_{c1}-\vx_1|| + (1+\epsilon)||\vx_2-\vx_{c2}|| \\
&\quad + (4+2\sqrt{\frac{N}{M}})\beta + \orderof{\delta} \\
&\leq (1+\epsilon)||\vx_1-\vx_2|| + (6+2\sqrt{\frac{N}{M}}+2\epsilon)\beta + \orderof{\delta} \ .
\end{split}
\end{equation}
We have shown the derivation of the upper bound. The derivation of the lower bound follows similar steps. We state the following lemma for the pairwise preserving property:
\begin{lemma}\label{lm:lipschitz_rip_ls}
Assume a Lipschitz continuous mapping $G(\cdot,\cdot): \R^D\times\R^{L-D}\rightarrow\R^N$ that has Lipschitz constant $T$ and satisfies: $||G(\vc, \vv_1) - G(\vc, \vv_2)|| \leq \beta,\ \forall \vc\in \setB^D(r_c), \vv_1, \vv_2 \in \setB^{L-D}(r_v)$, where $\setB^L(r)$ denotes a norm ball in $\R^L$ with radius $r$: $\setB^L(r) = \{ z | z\in\R^L, ||z||<r\}$ and $||\cdot||$ denotes the $L_2$ norm. Assume also a random matrix $\mPhi \in \R^{M \times N}$ whose entries are sampled from an i.i.d. Gaussian $\mathcal{N}(0, 1/M)$. Given some $0 <\epsilon<1$, $0<\delta<1$, if $M \geq \orderof{\frac{D}{\epsilon^2}\log\frac{Tr_c}{\delta}}$, then with probability at least $1-\delta$:
\begin{equation}
||\mPhi(\vx_1-\vx_2)|| \geq (1-\epsilon)||\vx_1-\vx_2|| - (6+2\sqrt{\frac{N}{M}} -2\epsilon)\beta - \orderof{\delta}
\end{equation}
and
\begin{equation}
||\mPhi(\vx_1-\vx_2)|| \leq (1+\epsilon)||\vx_1-\vx_2|| + (6+2\sqrt{\frac{N}{M}}+2\epsilon)\beta + \orderof{\delta}
\end{equation}
holds $\forall \vx_1,\vx_2 \in G(\setB^D(r_c), \setB^{L-D}(r_v))$.
\end{lemma}

Going from the pairwise distance preserving property to the recovery guarantee, we follow the similar steps as shown in deriving Lemma \ref{lm:recovery2}, which leads to the following lemma:
\begin{lemma} \label{lm:recovery1}
Assume a Lipschitz continuous mapping $G(\cdot,\cdot): \R^D\times\R^{L-D}\rightarrow\R^N$ that has Lipschitz constant $T$ and satisfies: $||G(\vc, \vv_1) - G(\vc, \vv_2)|| \leq \beta,\ \forall \vc \in \setB^D(r_c), \vv_1, \vv_2 \in \setB^{L-D}(r_v)$, where $\setB^L(r)$ denotes a norm ball in $\R^L$ with radius $r$: $\setB^L(r) = \{ z | z\in\R^L, ||z||<r\}$ and $||\cdot||$ denotes the $L_2$ norm. Assume also a random matrix $\mPhi \in \R^{M \times N}$ whose entries are sampled from an i.i.d. Gaussian $\mathcal{N}(0, 1/M)$. The compressed measurements are obtained as $\vy = \mPhi\vx^* + w$, where $\vx^*$ is the desired signal and $w$ is an independent additive noise. Define $\bar{\vx} = \argmin{\vx\in G(\setB^D(r_c), \setB^{L-D}(r_v))} ||\vx^* - \vx||$ and $\hat{\vx} = \argmin{\vx\in G(\setB^D(r_c), \setB^{L-D}(r_v))} ||\vy - \mPhi\vx||$. 

For some $0 < \epsilon< 1$, $0<\delta<\frac{1}{2}$, if $M \geq \orderof{\frac{D}{\epsilon^2}\log\frac{Tr_c}{\delta}}$, then with probability at least $1-\delta$:
\begin{equation}
||\vx^*-\hat{\vx}|| \leq \frac{5+2\sqrt{\frac{N}{M}}-\epsilon}{1-\epsilon}||\vx^*-\bar{\vx}|| + \frac{6+2\sqrt{\frac{N}{M}}-2\epsilon}{1-\epsilon}\beta + \frac{2}{1-\epsilon}||w|| + \orderof{\delta} \ .
\end{equation}
\end{lemma}

Lemma \ref{lm:recovery1} corresponds to the first part of our main theorem. The number of required measurements is linear with $D$ which is usually much smaller than the full latent variable dimension $L$. When the number of measurements increases, we can apply Lemma \ref{lm:recovery2} directly with $r = \sqrt{r_c^2+r_v^2}$ since $\setB^D(r_c)\times\setB^{L-D}(r_v) \subseteq \setB^{L}(\sqrt{r_c^2+r_v^2})$, and arrive at the second part of our main theorem. We have now finished the entire proof of Theorem \ref{thm:recovery}.

\twocolumn
\section{CONCLUSION}
This paper proposed an algorithm of using generative models to solve compressive sensing inverse problems. This method is fast to carry out, by exploiting a projector network, and is stable under high compression factor, by putting constraints on the latent variable space. It consistently produces high-quality images even when the observations are highly compressed. The proposed algorithm can be easily generalized to solve inverse imaging problems besides CS recovery. \footnote{The Tensorflow implementation of our tests can be found at https://github.com/sihan-zeng/f-csrg.}


\bibliographystyle{IEEEbib}
{\small\bibliography{references}}

\end{document}